\RequirePackage{fix-cm}
\documentclass[smallextended]{svjour3}       
\smartqed  
\usepackage{graphicx}
\usepackage[hyphens]{url}
\usepackage{times}
\usepackage{latexsym}
\usepackage{amsmath}
\usepackage{amssymb}
\usepackage{framed}
\usepackage{booktabs}
\usepackage{color, colortbl}
\usepackage{subfigure}
\usepackage{natbib}

\definecolor{light-gray}{gray}{0.70}

\newcounter{examplecounter}
\newenvironment{exampleb}{\begin{quote}
    \refstepcounter{examplecounter}
  \textbf{Example \arabic{examplecounter}} \\
}{
\end{quote}
}

\journalname{Language Resources and Evaluation}
\begin{document}

\title{An Annotated Corpus of Relational Strategies in Customer Service
\thanks{Under review in Language Resources and Evaluation}
}
%



\author{Ian Beaver         \and
        Cynthia Freeman	   \and
        Abdullah Mueen
}


\institute{Ian Beaver \at
			  Next IT Corporation \\
              Spokane Valley, WA USA \\
              (509) 242-0767 \\
              \email{ibeaver@nextit.com}
           \and
           Cynthia Freeman \at
			  Next IT Corporation \\
              Spokane Valley, WA USA \\
              (509) 242-0767 \\
              \email{cfreeman@nextit.com}
           \and
           Abdullah Mueen \at
			  Department of Computer Science \\
              University of New Mexico, USA \\
              (505) 277-1914 \\
              \email{mueen@unm.edu}
}

\date{Received: 16 August 2017 / Accepted: date}

\maketitle

\begin{abstract}
We create and release the first publicly available commercial customer service corpus with annotated \textit{relational segments}. Human-computer data from three live customer service Intelligent Virtual Agents (IVAs) in the domains of travel and telecommunications were collected, and reviewers marked all text that was deemed unnecessary to the determination of user intention. After merging the selections of multiple reviewers to create \textit{highlighted} texts, a second round of annotation was done to determine the classes of language present in the highlighted sections such as the presence of Greetings, Backstory, Justification, Gratitude, Rants, or Emotions. This resulting corpus is a valuable resource for improving the quality and relational abilities of IVAs. As well as discussing the corpus itself, we compare the usage of such language in human-human interactions on TripAdvisor forums. We show that removal of this language from task-based inputs has a positive effect on IVA understanding by both an increase in confidence and improvement in responses, demonstrating the need for automated methods of its discovery.
\end{abstract}

\section{Introduction}
\label{sec:intro}

Intelligent personal assistants such as Apple's Siri, Microsoft's Cortana, or Google Now are commonly used for answering questions and task optimization.  Many companies are deploying specialized automated assistants, known as Intelligent Virtual Agents (IVAs), for efficient problem resolution, cutting costs in call centers, and also as the first layer of technical and product support~\citep{icmi2013}.  In these business domains, IVA accuracy and efficiency directly impact customer satisfaction and company support costs.  In one case study~\citep{Caramenico2013}, a Fortune 50 insurance company experienced a $29\%$ reduction in contact center volume within five months of deploying an IVA on their website.  Domino's Pizza reported that product order time was reduced by $50\%$ through their IVA~\citep{Frearson2015}.

To better assist humans, IVA designers strive to support human-like interactions.  Take, for example, Amazon's Alexa Prize competition where student developers attempt to build IVAs that can carry on meaningful, coherent, and engaging conversations for 20 minutes~\citep{levy2016alexa}.  As IVAs become more human-like, we theorize that users will increasingly use \textbf{relational strategies} (e.g. self-exposure and justification) with IVAs similar to conversing with humans.  There is a large body of work on development of trust between humans engaged in virtual dialog~\citep{gibson2003virtual,ballantyne2004dialogue,holton2001building,coppola2004building}.  The focus of these works is on how relational strategies contribute to trust between human speakers.  From this literature, we predict the types of strategies humans may employ with IVAs as they relate to them in an increasingly human manner.

In customer service and personal assistant domains, trust is necessary between the human agent and customer.  The customer's issues must be viewed by the agent as legitimate for proper attention to be given.  Likewise, customers must trust that the agent is capable of assisting them and will not mistreat their information.  Current research shows that human-like virtual agents are associated with not only greater user trust but also trust resilience when the agent makes mistakes~\citep{de2016almost}.  To build trust with the agent, customers may establish credibility through small talk, self-exposure, and by providing justification of their requests~\citep{bickmore2001relational}.

In interactive question answering, such as dialogs with an IVA, understanding \textbf{user intent} is essential for the success of the IVA~\citep{chai2006towards}.  The intent can be defined as the interpretation of a user input that allows an agent to formulate the \textit{best} response.  However, when relational strategies are applied to IVAs, the additional language introduced is often unnecessary and can even obfuscate user intent.  Such language can lead to confusion in the IVA and a degradation of user experience in the form of clarification questions and wrong information.  

\begin{exampleb}
\label{ex1}
I need a ticket to Boston this Saturday, my son is graduating!
\end{exampleb}

In Example~\ref{ex1}, the fact that the customer's son is graduating is unnecessary for determining the user's intent to purchase a ticket.  By including unnecessary background information, the IVA may incorrectly deduce that the customer is booking a ticket \textit{for} his or her son instead. Thus, the identification of relational segments is a useful feature for an IVA; unfortunately, no corpus of annotated relational segments exists to develop identification techniques~\citep{serban2015survey}.

This lack inspired us to create such a corpus. Within this corpus, we needed to not only identify the location of relational language but also label its type (Gratitude, Greetings, etc.) so that automated methods to determine the relational strategy in use can be explored.

If these strategies are practiced by users of IVAs, it is important to identify them; enabling IVAs to separate such language can help better clarify the users' main intention.  For IVAs to become more human-like, determining which segments of a request are relational is necessary to allow these IVAs to both understand the user intent correctly and to include empathetic or reciprocal relational strategies.

The identification of relational strategies in a single conversational turn can be structured as a multi-intent detection problem.  The user not only wants the task completed (the \textit{primary} intent); they may also attempt to build credibility or some common ground with the IVA (the \textit{secondary} intent).  Segments of text such as justification or backstory can be annotated as secondary intent and ignored while determining the primary intent.  Once relational language is isolated, a separate classification can determine what relational strategies are in use and how to properly respond.

Multi-intent detection within dialog systems is still an emerging field; in recent work, only one intent is assumed to be present per turn~\citep{sarikaya2016overview}. A few methods exist such as~\cite{xu2013exploiting} which uses multi-label learning and~\cite{kim2016two} which employs a two-stage intent detection strategy.  However,~\cite{xu2013exploiting} provided no explanation of how data was annotated nor any mention of annotator agreement. In~\cite{kim2016two}, multi-intent data was fabricated by concatenating all combinations of single-intent sentences.

In this article, we provide several contributions. Most importantly, we create the first publicly available customer service corpus with annotated relational segments.  We propose an evaluation measure and set a baseline by comprehensive human annotation, ultimately confirming that the addition of relational language can obfuscate the user's intention to IVAs not designed to recognize it. Along with annotated relational segments, our corpus includes multi-intent requests to further research in multi-intent detection.  We analyze human agreement in determining the presence of multiple intents so that future research on multi-intent detection can be evaluated in the light of prior human performance.  Through these contributions, we hope to encourage further research and ultimately aid in the design of more intelligent IVAs.

In the following sections, we discuss in detail how the data was collected, annotated, and merged to create \textbf{highlighted} sections. Another round of review was then done on these highlighted sections to determine the class of language present in these sections (e.g. Greeting, Gratitude, etc). We then measure and compare the frequency of relational strategies when users present their requests to IVAs versus humans.  Finally, we conduct an experiment with three commercial IVAs, demonstrating that removal of relational strategies lowers confusion and leads to improved responses.

\section{Data Collection}
\label{sec:data}

Next IT Corporation designs and builds IVAs on behalf of other companies and organizations, typically for customer service automation.  This unique position allows access to a large number of IVA-human conversations that vary widely in scope and language domain.  We selected IVAs for data collection based on the volume of conversations engaged in, the scope of knowledge, and the diversity of the customer base.

For diversity, we considered whether the target user base of the IVA was local, regional, national, or international and mapped the locations of the users engaging in conversations to visually verify.  We only considered IVAs that had a national or international target user base and did not appear to have a dominate regional clustering to ensure that conversations were well distributed across users from different regions.  This was to control for relational styles that may differ between regions.

IVAs deployed in domains that were highly sensitive, such as human resource management or health care, were not considered.  As a result, human-computer data was collected from three live customer service IVAs in the language domains of airline, train travel, and telecommunications.  Each agent met our criteria of a broad knowledge base, sufficient conversation volume, and a very diverse user base.

The selected IVAs are implemented as mixed-initiative dialog systems, each understanding more than 1,000 unique user intentions.  The IVAs have conversational interfaces exposed through company websites and mobile applications.  In addition, the IVAs are multi-modal, accepting both speech and textual inputs, and also have human-like qualities with simulated personalities and interests.  A random sample of 2,000 conversations was taken from each domain.  The samples originate from conversation logs during November 2015 for telecommunications and train travel and March 2013 for airline travel.  There were 127,379 conversations available in the logs for the airline IVA.  The telecommunications and train travel logs contained 837,370 and 694,764 conversations, respectively.  The first user turn containing the problem statement was extracted.  We focus on the initial turn as a user's first impression of an IVA is formed by its ability to respond accurately to his or her problem statement, and these impressions persist once formed~\citep{bentleyU,madhavan2006automation}.  Therefore, it is imperative that any relational language present does not interfere with the IVA's understanding of the problem statement.

Finding a large mixed-initiative human-human customer service dataset for comparison with our human-computer dialogs proved difficult.  Despite mentions of suitable data in~\cite{vinyals2015neural} and~\cite{roy2016qart}, the authors did not release their data.  Inspecting the human-human chat corpora among those surveyed by \cite{serban2015survey} revealed only one candidate: the Ubuntu Dialogue Corpus~\citep{lowe2017training}.  The corpus originates from an Internet Relay Chat (IRC) channel where many users discuss issues relating to the Ubuntu operating system.  After a user posts a query on the channel, all following threads between the querying user and each responding user are isolated to create two-way task-specific dialogs.  However, we want to study the initial problem statements to compare their composition with those extracted from our data.  In the Ubuntu corpora, these are posed to a large unpaid audience in the hopes that someone will respond.  The observed relational language and behavior was, therefore, no different than problem statements inspected in other IRC or forum datasets, and, for our purposes, was no more fitting than any other forum or open IRC dataset.

In addition, we desire to not just measure relational language content but also feed the problem statements into an IVA and measure the effect of any relational language on its understanding of the user intent.  To do this, we needed requests that were very similar to those already handled by one of the selected IVAs to have any hope of the user intent already existing in the agent's knowledge base.  Unsatisfied with the Ubuntu dataset, we instead focused on publicly visible question and answering data in domains similar to those of the selected IVAs.

Upon searching publicly visible chat rooms and forums in the domains of travel and telecommunications support, we found the TripAdvisor.com airline forum to be the closest in topic coverage.  This forum includes discussions of airlines and polices, flight pricing and comparisons, flight booking websites, airports, and general flying tips and suggestions.  We observed that the intentions of requests posted by users were very similar to that of requests handled by our airline travel IVA.  While a forum setting is a different type of interaction than chatting with a customer service representative (user behavior is expected to differ when the audience is not paid to respond), it was the best fit that we could obtain for our study and subsequent release.  A random sample of 2,000 threads from the 62,736 present during August 2016 was taken, and the initial post containing the problem statement was extracted.  We use \textbf{request} hereafter to refer to the complete text of an initial turn or post extracted as described.

\subsection{Annotation}
\label{subsec:annotations}
From our four datasets of 2,000 requests each, we formed two equally-sized partitions of 4,000 requests with 1,000 pulled from every dataset. Each partition was assigned to four reviewers; thus, all 8,000 requests had exactly four independent annotations.  All eight reviewers were employees of Next IT Corporation who volunteered to do the task in their personal time.  As payment, each reviewer received a \$150 gift card.

\begin{table}[b]
\caption{Dataset statistics. The Multi-Intent column represents the count of Requests where one or more reviewers flagged it as containing more than one user intention.  The Unnecessary column represents the percentage of Single Intent requests where one or more reviewers selected \textit{any} text as being unnecessary in determining user intent.  Avg. Length is the number of words present in All Requests, on average.}
\label{tbl:stats}
\begin{center}
 \begin{tabular}{ l ccccc} 
 \toprule
  & All Requests & Multi-Intent & Single Intent & Unnecessary & Avg. Length \\
 \cmidrule(l){2-6}
 \textbf{TripAdvisor} & 2000 & 734 & 1266 & 94.1\% & 93.26  \\

 \textbf{Telecom} & 2000 & 149 & 1851 & 77.3\% & 19.81 \\

 \textbf{Airline} & 2000 & 157 & 1843 & 68.6\% & 21.64 \\

 \textbf{Train} & 2000 & 201 & 1799 & 55.3\% & 20.07 \\
 \bottomrule
\end{tabular}
\end{center}
\end{table}

The reviewers were instructed to read each request and mark \textit{all} text that appeared to be additional to the user intention.  The reviewers were given very detailed instructions, shown in Appendix B, and were required to complete a tutorial demonstrating different types of relational language use before working on the actual dataset.  As the data was to be publicly released, we ensured that the task was clear.  If more than one user intention was observed, the reviewer was instructed to flag it for removal.  This was a design decision to simplify the problem of determining language necessary for identifying the user intention.  Furthermore, as mentioned in section~\ref{sec:intro}, IVAs with the ability to respond to multiple intentions are not yet commonplace. Although flagged requests were not used for further analysis, they are included in the released data to enable future research.  After discarding all multi-intent requests, 6,759 requests remained.  Per-dataset statistics are given in Table~\ref{tbl:stats}.

A request from the TripAdvisor data is given in Example~\ref{ex2} below.  A reviewer first read over the request and determined that the user intent was to gather suggestions on things to do during a long layover in Atlanta.  The reviewer then selected all of the text that they felt was not required to determine that intent.  This unnecessary text in Example~\ref{ex2} is shown in gray.  Each of the four reviewers performed this task independently, and we discuss in the next sections how we compare their agreement and merged the annotations.

\begin{exampleb}
\label{ex2}
{\small
\noindent \textbf{Original Request:} Hi My daughter and I will have a 14 hour stopover from 20.20 on Sunday 7th August to 10.50 on Monday 8th August. Never been to Atlanta before. Any suggestions? Seems a very long time to be doing nothing. Thanks

\vspace{0.3cm}

\noindent \textbf{Determine User Intent:} \textit{Things to do during layover in Atlanta}

\vspace{0.3cm}

\noindent \textbf{Annotated Request:} \textcolor{light-gray}{Hi} My daughter and I will have a 14 hour stopover \textcolor{light-gray}{from 20.20 on Sunday 7th August to 10.50 on Monday 8th August.} Never been to Atlanta before. Any suggestions? \textcolor{light-gray}{Seems a very long time to be doing nothing. Thanks}
}
\end{exampleb}

Reviewers averaged 1 request per minute over 1,000 requests on TripAdvisor data and 4 per minute over 3,000 requests from the three IVA datasets.  We observed that each of the eight reviewers required 29 hours on average to complete their 4,000 assigned requests.

\section{Annotation Alignment}

To compare the raw agreement of annotations between two reviewers, we use a modification of \textit{alignment} scores, a concept in speech recognition from hypothesis alignment to a reference transcript~\citep{zechner2000minimizing}. We modify this procedure as insertions and deletions do not occur. Reviewers mark sequences of text as being unnecessary in determining user intention. When comparing annotations between two reviewers, an error ($e_i$) is considered to be any character position $i$ in the text where this binary determination does not match between them.  The alignment score can be calculated as:

 $$align = \frac{n - \sum_{i=1}^{n} e_i}{n}$$ 

\noindent where $n$ is the total number of characters.  Thus, $align \in [0,1]$ where $1$ is perfect alignment.  Reviewers may or may not include whitespace and punctuation on the boundaries of their selections which can lead to variations in $e_i$.  Therefore, when two selections overlap, we ignore such characters on the boundaries while determining $e_i$.  Figure~\ref{fig:align} shows a fabricated example of alignment between two annotations.  In the first selection, the trailing whitespace and punctuation are ignored as they occur within overlapping selections.  Notice, however, that whitespace and punctuation count in the last selections as there is no overlapping selection with the other reviewer; therefore, there is no possibility of disagreement on the boundaries.

\begin{figure}[h]
\definecolor{shadecolor}{rgb}{0.96,0.96,0.96}
\begin{snugshade}
$A$: [Hi, ]I need a new credit card[\underline{, my old doesn't work any more.}] Can you help? \\
$B$: [Hi], I need a new credit card, my old doesn't work any more.[\underline{ Can you help?}] \\

{\small
$n = 73$ \hspace{0.45cm} $\sum_{i=1}^{73} e_i = 45$ \hspace{0.45cm} $align_{AB} = \frac{73 - 45}{73} = 0.384$ }
\end{snugshade}
\caption{Example alignment scoring between two fabricated annotations $A$ and $B$. Text between ``['' and ``]'' was marked as unnecessary for intent determination. Positions with an alignment error are underlined.}
\label{fig:align}
\end{figure}

\begin{figure}[h]
\centering\subfigure[Overall alignment scores]{\includegraphics[scale=.7]{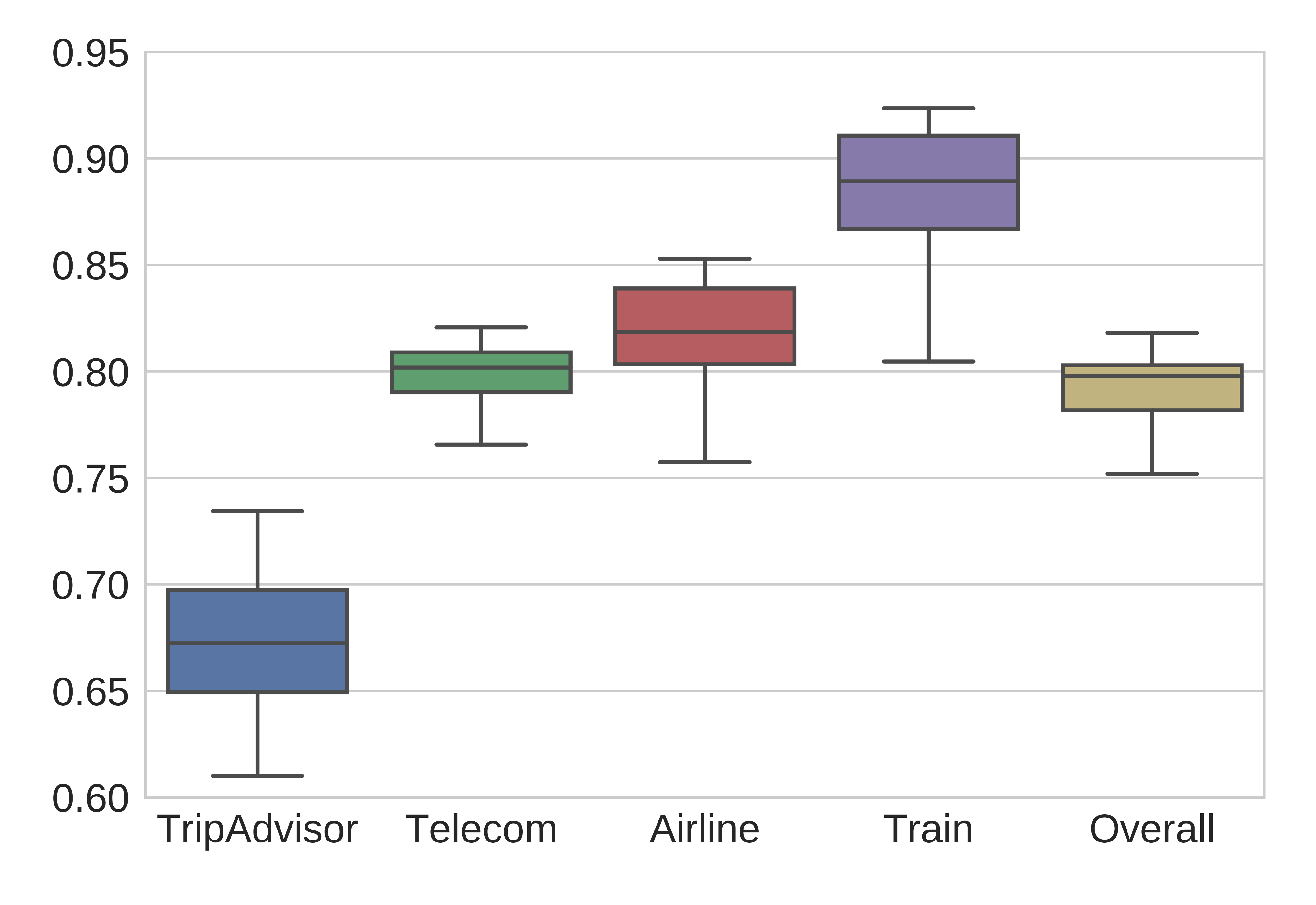}}
\subfigure[Alignment scores when reviewers agree that additional language is present]{\includegraphics[scale=.7]{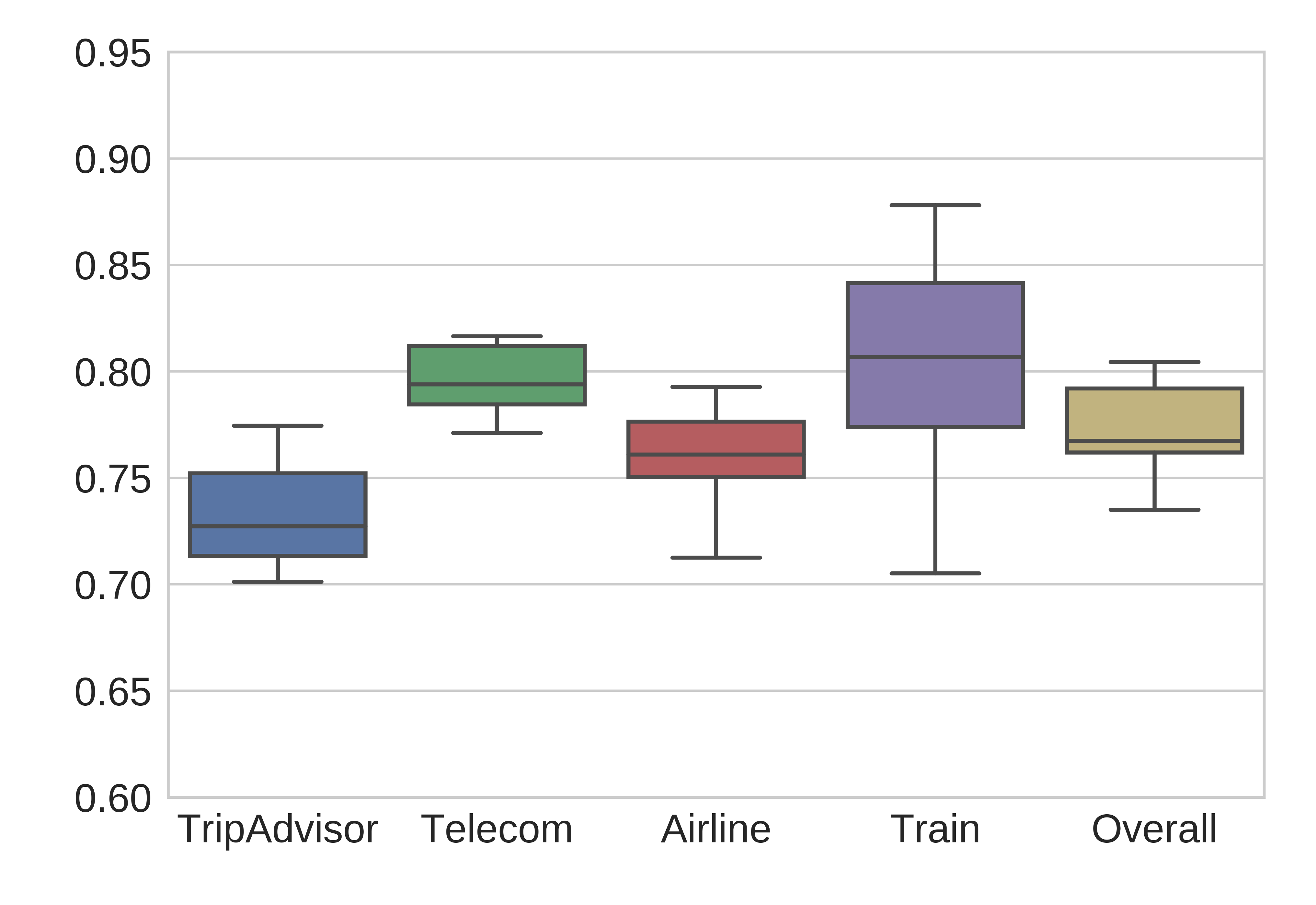}}
\caption{The distribution of average alignment scores between all four annotations per dataset is shown in \textbf{(a)}. We compute average alignment scores where all reviewers agree that additional language is present in \textbf{(b)}.}
\label{fig:alignments12}
\end{figure}

The alignment score was calculated for every request between all four annotations and then averaged.  For example, an alignment score was calculated for each request between reviewer $A$ and $B$, $A$ and $C$, $A$ and $D$. The same process was repeated between reviewer $B$ and $C$, $B$ and $D$, then $C$ and $D$.  Finally, alignment scores between all unique pairs of reviewers over all requests were averaged per dataset.  The distribution of average scores per dataset is shown in Figure~\ref{fig:alignments12} \textbf{(a)}.  It may appear, at first, that two annotators could inflate the dataset alignment score by simply making annotations infrequently.  However, as each request had four annotators, the average alignment score would actually be lower as those reviewers would have large error compared to the other two.  The per dataset alignment averages can, in fact, be higher if a dataset has a large number of requests where \textit{no} reviewer selected any text.

Therefore, it is interesting to remove the effect of these cases and compare the ability of reviewers to agree on the selection boundaries given they both agree that selection is necessary.  To measure this, we compute average alignment scores where both reviewers agree that additional language is present, shown in Figure~\ref{fig:alignments12} \textbf{(b)}.  Observe that although the Train dataset has the highest overall alignment in both cases, it is lower when the reviewers both select text, indicating it has many cases where no reviewers selected anything (which is in agreement with Table~\ref{tbl:stats}).  In the case of TripAdvisor, it appears that there are a significant number of requests where one or more reviewers do not select text, but the others do, lowering the overall alignment score in Figure~\ref{fig:alignments12} \textbf{(a)}.

Alignment based on word-level instead of character-level agreement was also considered.  For each word, if the reviewer selected at least 50\% of the word it was considered to be marked.  This resolves situations where a reviewer accidentally missed the first or last few characters of a word in their selection.  However, this may introduce errors where two letter words have only one character selected.  In this case it is impossible to automatically decide if the reviewer meant to select the word or not as always selecting such words will be susceptible to the same error.  Besides this ambiguous case, we felt it was safe to assume that words of longer length with less than half of the word selected were not intended to be marked.

Selected words were then used in place of selected characters in calculating the alignment scores between the reviewers in the same manner as Figure~\ref{fig:align}.  We discovered that the alignment scores were only 0.2\% different on average across the datasets than the character level alignment scores shown in Figure~\ref{fig:alignments12}.  This indicates that reviewers are rarely selecting partial words, and any disagreement is over \textit{which} words to include in the selections.  Therefore, in the released corpus and in this article, we consider selections using absolute character position which retains the reviewers' original selection boundaries.

\subsection{Agreement Between Reviewers}
\label{subsec:agree}

\begin{table}
\caption{Reviewer agreement on if any text should be selected.  For example, row 3 is the number of requests with selections by at least three reviewers.}
\label{tbl:selectkappa}
\begin{center}
\begin{tabular}{cllll}
\toprule
& TripAdvisor & Train & Airline & Telecom \\
\cmidrule(l){2-5}
$\kappa$ & 0.270 & 0.450 & 0.405 & 0.383 \\
 \cmidrule(l){2-5}
1 & 1192 & 995 & 1264 & 1431 \\
2 & 1092 & 709 & 948 & 1154 \\
3 & 863 & 458 & 644 & 795 \\
4 & 534 & 205 & 292 & 410 \\
\bottomrule
\end{tabular}
\end{center}
\end{table}

As it is difficult to determine how often all reviewers agree additional language is present from alignment scores alone, we measured reviewer agreement on the presence of additional language and multiple user intentions.  For additional language presence, we calculated Fleiss' $\kappa$ over the annotations where the classes compared were if a reviewer did or did not select text. As demonstrated in Table~\ref{tbl:selectkappa}, regardless of domain, this is a subjective task.  While there is moderate agreement in the Train and Airline sets, the TripAdvisor set, in particular, is lower in agreement which reinforces our previous observations in Figures~\ref{fig:alignments12} \textbf{(a)} and \textbf{(b)}.  Due to the sensitivity of $\kappa$ measurements~\citep{feinstein1990high,guggenmoos1993reliable}, these values must be interpreted in light of the task.  Despite the lower values, we are only measuring presence or absence of unnecessary language, and these two categories did not necessarily occur in equal frequencies.  Under these conditions, according to~\cite{bakeman1997detecting}, a $\kappa$ between 0.2 and 0.45 may suggest reviewer reliabilities between 80\% to 90\%, respectively.  Therefore, despite the lower values for $\kappa$, the individual reviewer annotations appear reliable and can be further improved when merged based on agreement as discussed in the following section.

\begin{exampleb}
\label{ex:3}
\noindent $R1$: Our tv reception is horrible. \textcolor{light-gray}{is there an outage in my area?}

\vspace{0.1cm}
\noindent $R7$: \textcolor{light-gray}{Our tv reception is horrible.} is there an outage in my area?
\end{exampleb}

We did observe situations where two reviewers disagree on the real intent of the user, therefore,  causing conflict in the selection of unnecessary text.  While these were rare, Example~\ref{ex:3} demonstrates how even humans sometimes struggle with determining the intention of written requests.  Reviewer 1 appears to believe that the primary intent of the user is to notify the agent about poor television reception, and the query about the outage in the area is out of curiosity.  However, reviewer 7 appears to believe the primary intent is to discover if a cable outage is present in the area, and the complaint about reception justifies the query.  The effects of these disagreements on intent can be mitigated by merging the annotations based on the number of reviewers who agreed on a selected character.

Next, we considered the reviewers' determination of multiple intentions.  A $\kappa$ was calculated over how reviewers flagged requests containing more than one user intention.  As shown in Table~\ref{tbl:multikappa}, we see somewhat similar performance in this task as in the previous selection task.  This table demonstrates the difficulty of multi-intent detection, even for humans.  The domain does not seem to be a factor as $\kappa$ is similar across datasets.  It is apparent, however, that in the forum setting, users are much more likely to insert multiple intentions in a single request than in a chat setting.

\begin{table}
\caption{Reviewer agreement on multi-intent detection.  For example, row 3 is the number of requests flagged as containing multiple intentions by at least three reviewers.}
\label{tbl:multikappa}
\begin{center}
\begin{tabular}{l p{1.8cm} p{0.9cm} p{1.0cm} p{1.2cm}}
\toprule
 & TripAdvisor & Train & Airline & Telecom \\
 \cmidrule(l){2-5}
$\kappa$ & 0.415 & 0.374 & 0.434 & 0.386 \\
 \cmidrule(l){2-5}
1 & 734 & 201 & 157 & 149 \\
2 & 480 & 85 & 69 & 56 \\
3 & 275 & 50 & 38 & 32 \\
4 & 71 & 8 & 15 & 11 \\
\bottomrule
\end{tabular}
\end{center}
\end{table}

\begin{figure}
\subfigure[Alignment between group 1 reviewers.]{\includegraphics[scale=.46]{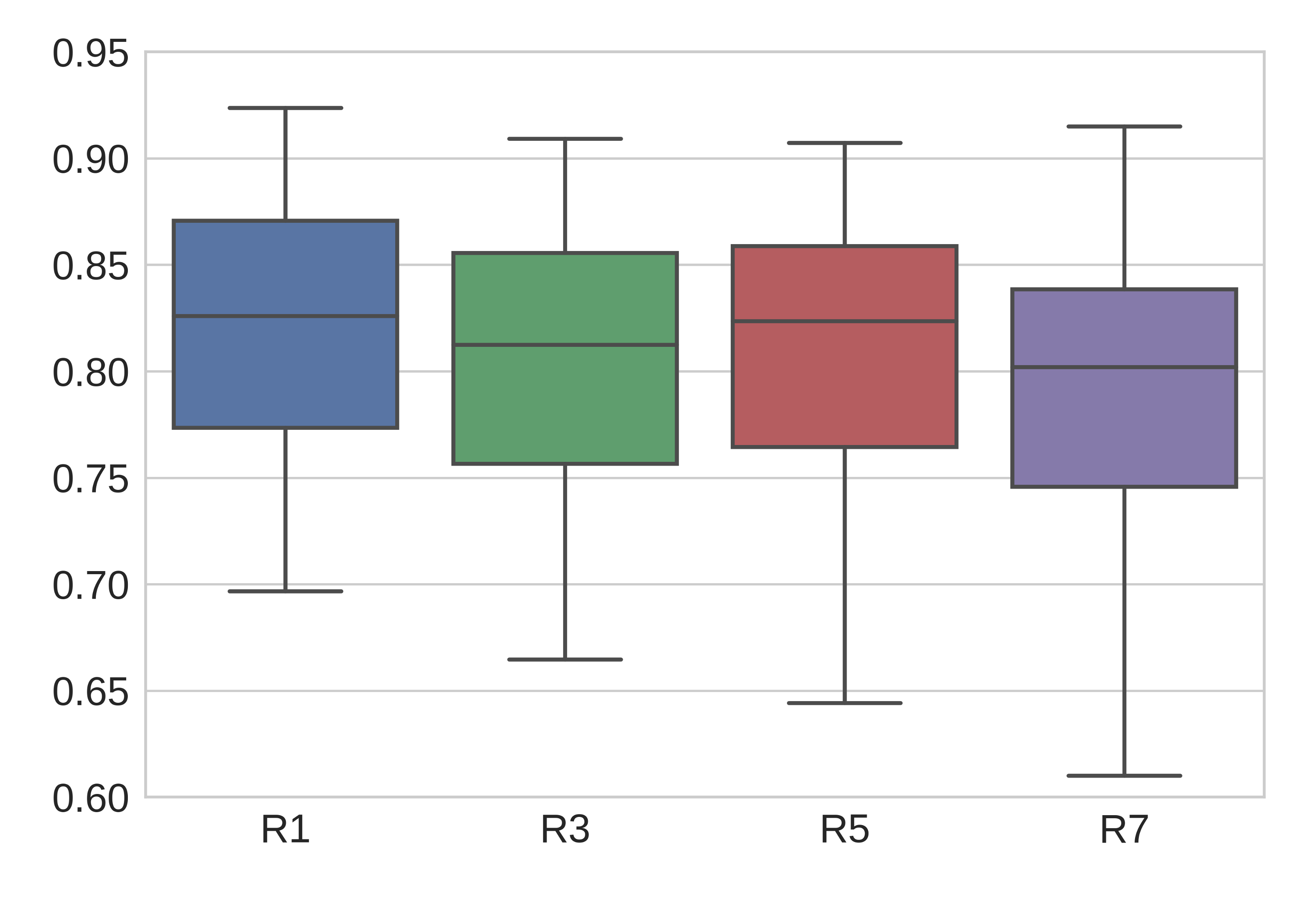}}
\hfill
\subfigure[Alignment between group 2 reviewers.]{\includegraphics[scale=.46]{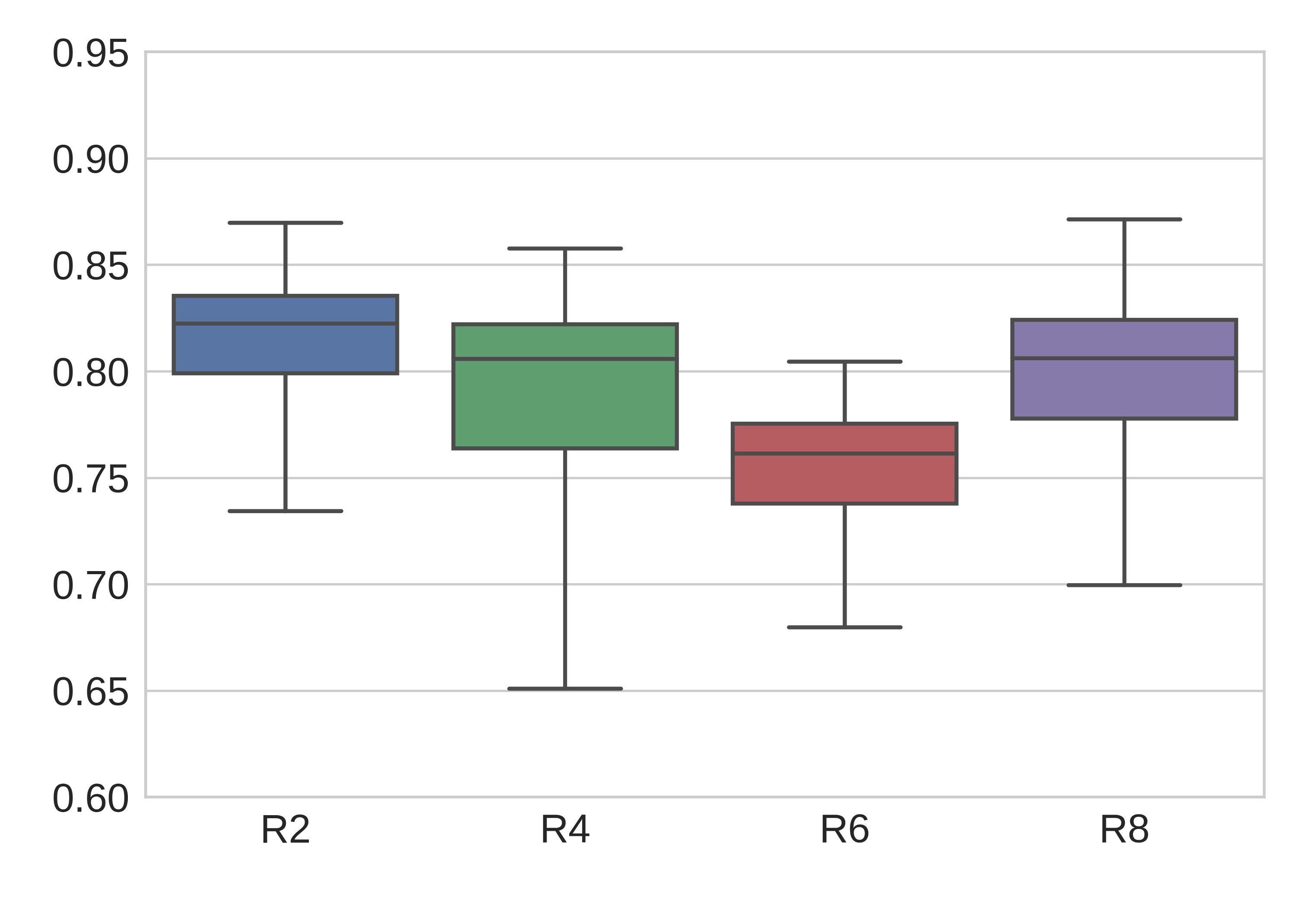}}
\caption{Alignment scores between each reviewer and the other three members of their group, averaged across the four datasets.}
\label{fig:align_p1_p2}
\end{figure}

How reviewers compare to the rest in their selections is another aspect to be considered.  Figure~\ref{fig:align_p1_p2} \textbf{(a)} compares how each reviewer agreed with the other 3 in the first group. We can see that, overall, the mean is very close. However, reviewer 7, in particular, had more variation in his or her selections. Similarly, Figure~\ref{fig:align_p1_p2} \textbf{(b)} compares how each reviewer agreed with the other 3 in the second group.  In the second group, we see slightly more disagreement, particularly with reviewer 6.  This could be because he did not interpret the user intention the same as others or because the reviewer was more generous or conservative in selections compared to the others in the group.

\subsection{Merging Selections By Agreement}
\label{sec:merge}

\begin{figure}[ht]
\includegraphics{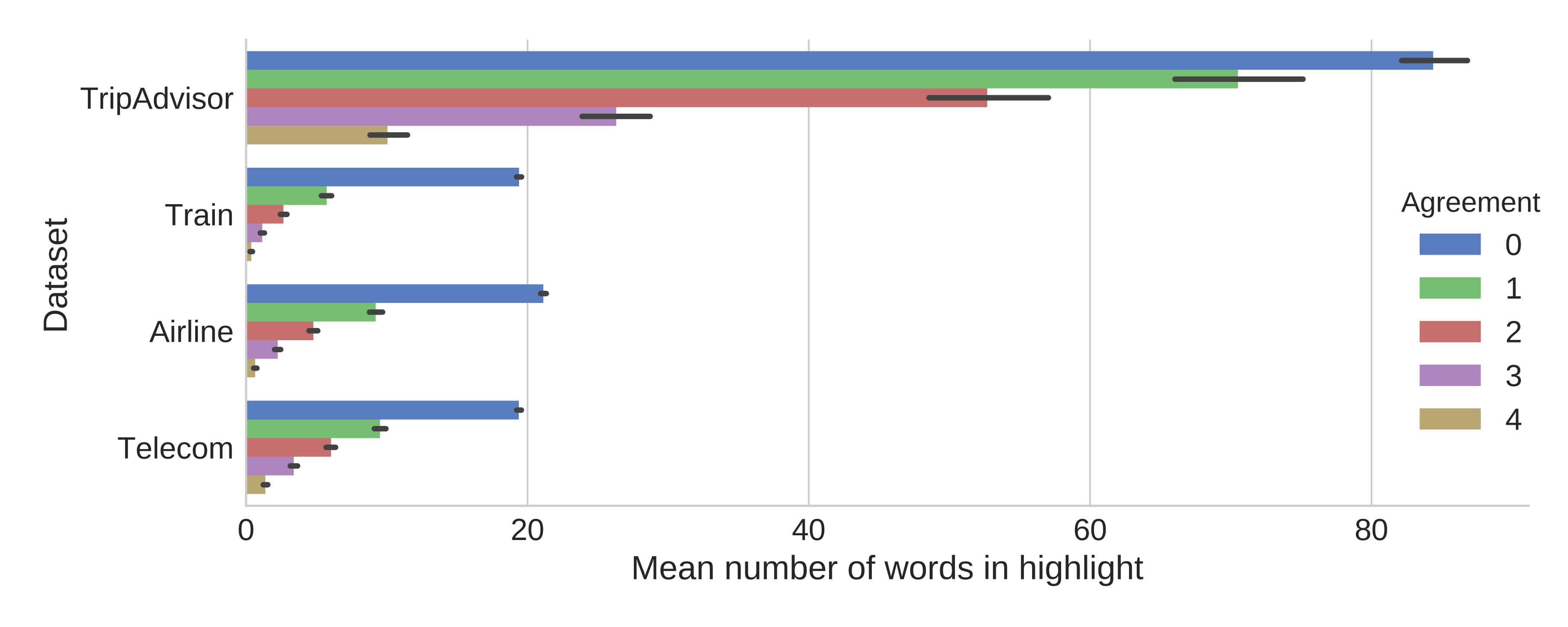}
\caption{Mean number of words highlighted per request by dataset.  Agreement is the number of reviewers who marked the same word for removal, where 0 is the original request length.}
\label{fig:numselection}
\end{figure}

The four annotations per request were \textbf{merged} using the following strategy: for every character position in the request, if at least a threshold of two annotations contained that position, \textbf{highlight} it.  To quantify the average reduction of request size, we count the number of words highlighted for each level of reviewer agreement.  In Figure~\ref{fig:numselection}, we can see that as the agreement required increases, the size of the highlight decreases significantly.

\section{Annotating Relational Content}
\label{sec:annotate-rc}
To determine the use of relational strategies, a second round of manual analysis was performed. An increase in agreement corresponds to a significant removal of remaining annotations as can be seen in Figure~\ref{fig:numselection}.  Therefore, in order to have sufficient data for analysis given the sample size, an agreement of two is used.  A comparison of relational annotation using all agreement levels is left for future works.

\begin{figure}
\includegraphics[scale=0.5]{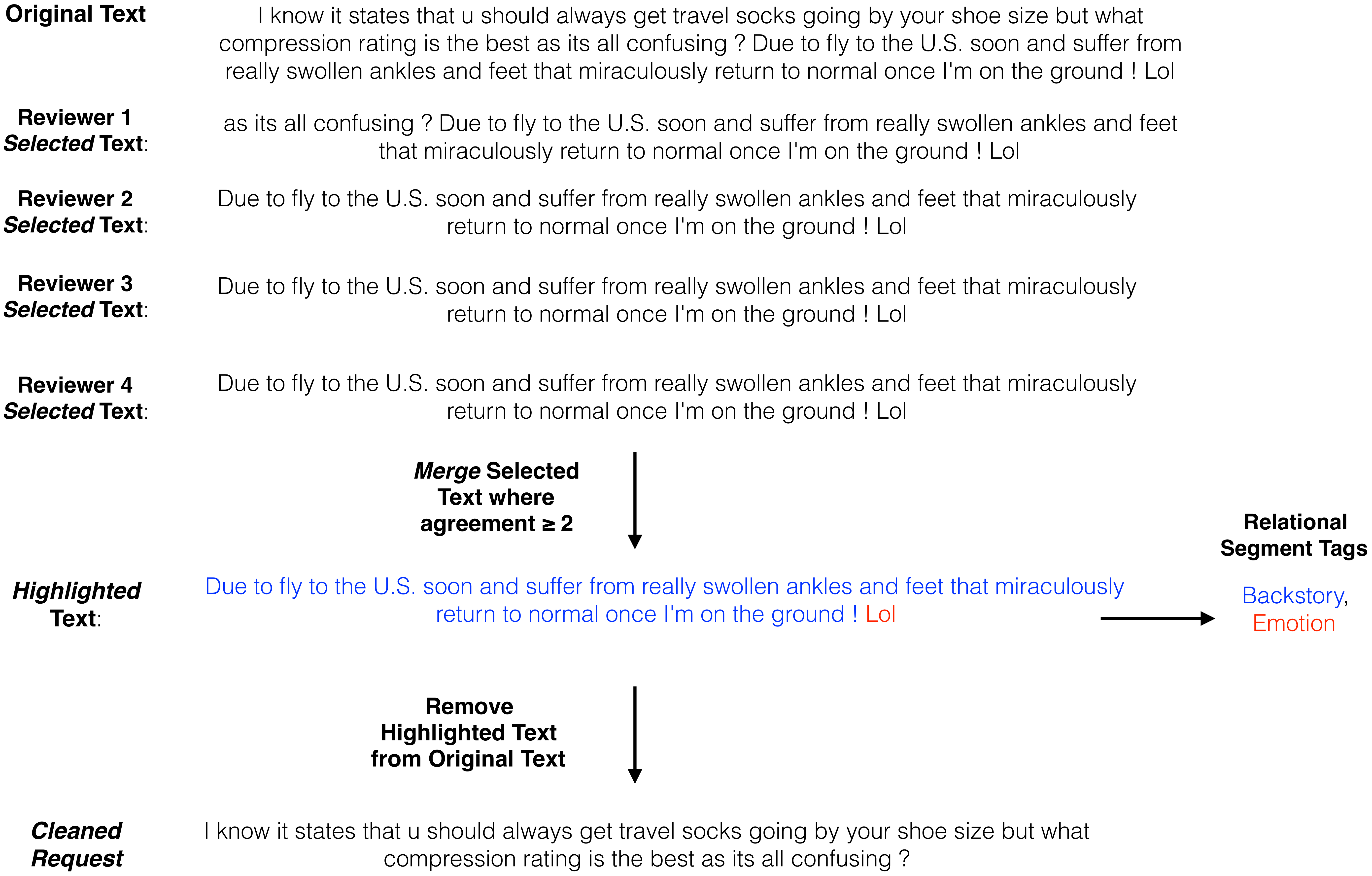}
\caption{An overview of the review and merging process.  In this example from the TripAdvisor corpus, reviewers 2, 3, and 4 all agree on which text is unnecessary. \textit{Selections} are \textit{merged} to form \textit{highlighted} text that is then removed from the original text to create a \textit{cleaned request}. A second round of annotation was done on highlighted texts to determine the classes of language present. The colors of the text correspond to the class present.}
\label{fig:process}
\end{figure}

Once merged, highlighted sections were analyzed by the authors to determine the classes of language present.  Each such section was evaluated and given one or more of the following tags: \textit{Greeting, Backstory, Justification, Gratitude, Rant, Express Emotion, Other}. See Figure~\ref{fig:process} for an overview of the entire process.

\textbf{Greetings} are a common relational strategy humans use to build rapport with other humans and machines~\citep{lee2010receptionist}.

\textbf{Backstory} is a method of self-exposure that may be employed by the customer.  In Example~\ref{ex1} given in section~\ref{sec:intro}, the customer included the fact that he or she is attending a graduation as a means of self-exposure.  This may be an attempt to build common ground with the agent or it may indicate the importance of the trip and motivate the agent to help the customer succeed.  

\textbf{Justification} is used by the customer to argue why the agent should take some action on the part of the customer.  For instance, when trying to replace a defective product, a customer may explain how the product failed to establish credibility that the product was at fault.

\textbf{Gratitude}, like greetings, are used by humans to also build rapport with humans and machines~\citep{lee2010receptionist}.

\textbf{Ranting} is a means of expressing dissatisfaction when a customer feels frustrated, ignored, or misunderstood. In computer-mediated conversations, the non-verbal emotional cues present in face-to-face conversations are missing; thus, humans resort to such negative strategies to convey their emotions~\citep{laflen2012okay}. For tagging purposes, we define a \textit{Rant} to encompass any excessive complaining or negative narrative.

\textbf{Expressing emotions} can be a means of showing displeasure when a customer feels a conversation is not making adequate progress or in reaction to an unexpected or disagreeable agent response.  This can also indicate joking or other positive emotional expression. The tag \textit{Express Emotion} is used as a catch-all for any emotional statement that is not covered by \textit{Rant}.  Examples would be: \textit{``i love that!", ``UGH!", ``WHY???"}.

The \textbf{Other} tag indicates that some or all of the section does not contain any relational language.  This is commonly a restatement of the primary intent or facts that reviewers marked as unnecessary.

\subsection{Analysis of Relational Tags}

\begin{figure}[h]
\centering\includegraphics{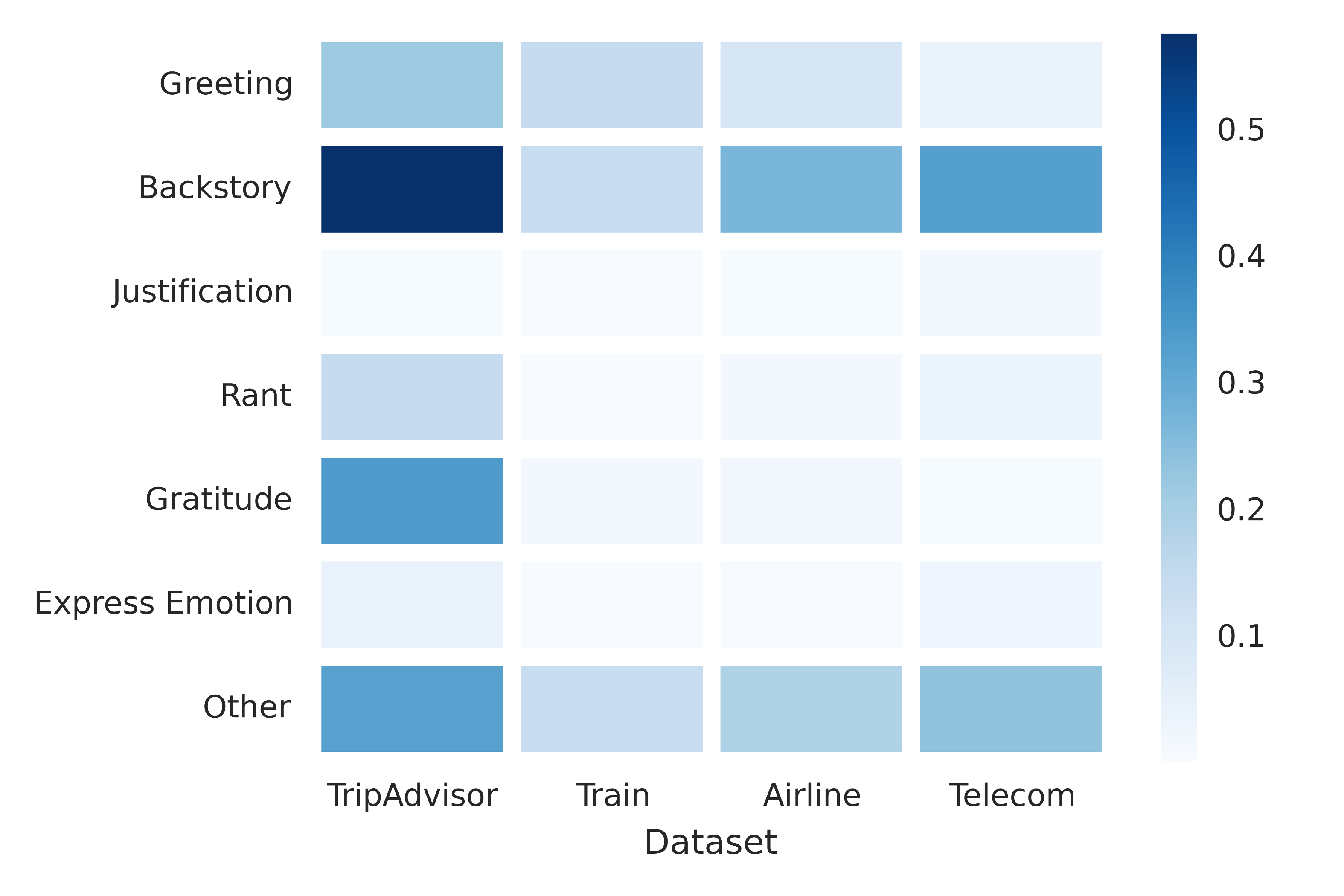}
\caption{Incidence of relational language per dataset.  An incidence of 0.5 means the tag is present in 50\% of all requests.}
\label{fig:tags_hm2}
\end{figure}

As shown in Figure~\ref{fig:tags_hm2}, we see that backstory is more common in human-to-human forum posts.  However, both Airline and Telecom IVAs also have a significant amount of backstory.  Although minimal, ranting and justification were present in Telecom.  The Train dataset appeared to contain the least amount of relational language.  It is difficult to speculate why without deeper analysis of the user demographic, the presentation of the IVA on the website, and the IVA knowledge base.

\begin{figure}[h]
\centering\includegraphics{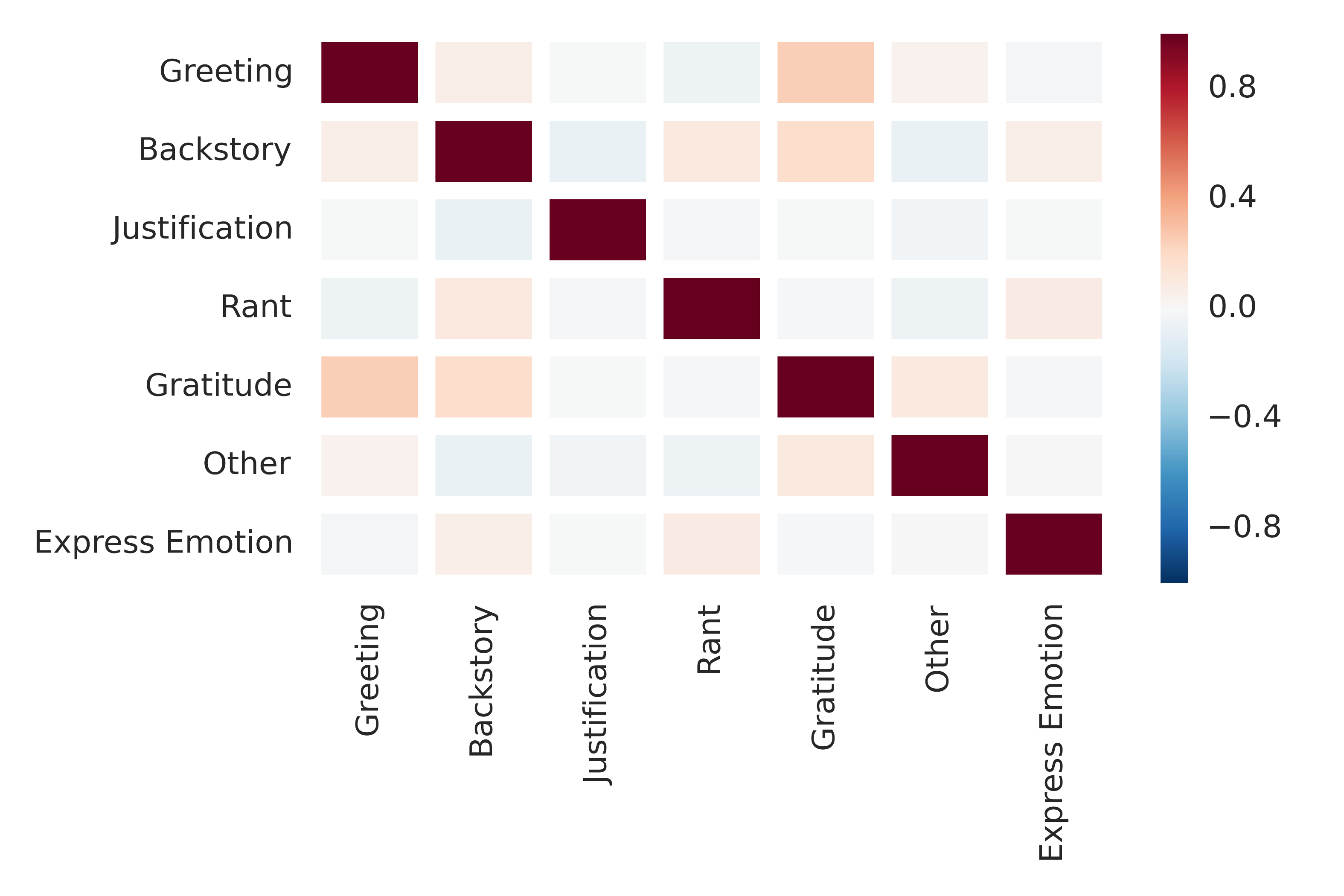}
\caption{Pearson coefficients of tag correlation across datasets.}
\label{fig:tags_hm}
\end{figure}

The correlation between tags is shown in Figure~\ref{fig:tags_hm}.  When greetings are present, it appears that there is a likelihood there will also be gratitude expressed which agrees with the findings in~\cite{lee2010receptionist} and~\cite{makatchev2009relating}.  Also interesting is the apparent correlation between backstory and gratitude.  Those that give background on themselves and their situations appear more likely to thank the listener.  Ranting appears to be slightly negatively correlated with greetings, which is understandable assuming frustrated individuals are not as interested in building rapport as they are venting their frustrations.

\section{Experiments and Results}

To measure the effect on IVA performance and determine what level of reviewer agreement is acceptable, we first constructed highlights for the 6,759 requests using all four levels of reviewer agreement. Next, four \textit{cleaned} requests were generated from each original request by removing the highlighted portion for each level of agreement resulting in 27,036 requests with various amounts of relational language removed.

Every unaltered request was fed through its originating IVA, and the intent confidence score and response was recorded.  We then fed each of the four cleaned requests to the IVA and recorded the confidence and response.  The TripAdvisor data was fed through the Airline IVA as it provided the most similar domain.  This was also a test to see if lengthy human-to-human forum posts could be condensed and fed into an existing IVA to generate acceptable responses.  We observed an increase in confidence in all domains with an average of 4.1\%.  The Telecom set, which had the highest incidence of backstory outside of TripAdvisor, gained 5.8\%.

In addition to intent confidence, we measured the effect of relational language removal on overall IVA understanding.  An A-B test was conducted where four reviewers were shown the user's original request along with the IVA response from the original request and the IVA response from a cleaned request.  They were asked to determine which, if any, response they believed better addressed the request. If the original IVA response was preferred, it was assigned the value -1. If the response to the cleaned request was preferred, it was assigned the value 1. Finally, if neither response even remotely addressed the user's request or if both responses were comparable, it was given the value 0.

\begin{figure}
\includegraphics[scale=.68]{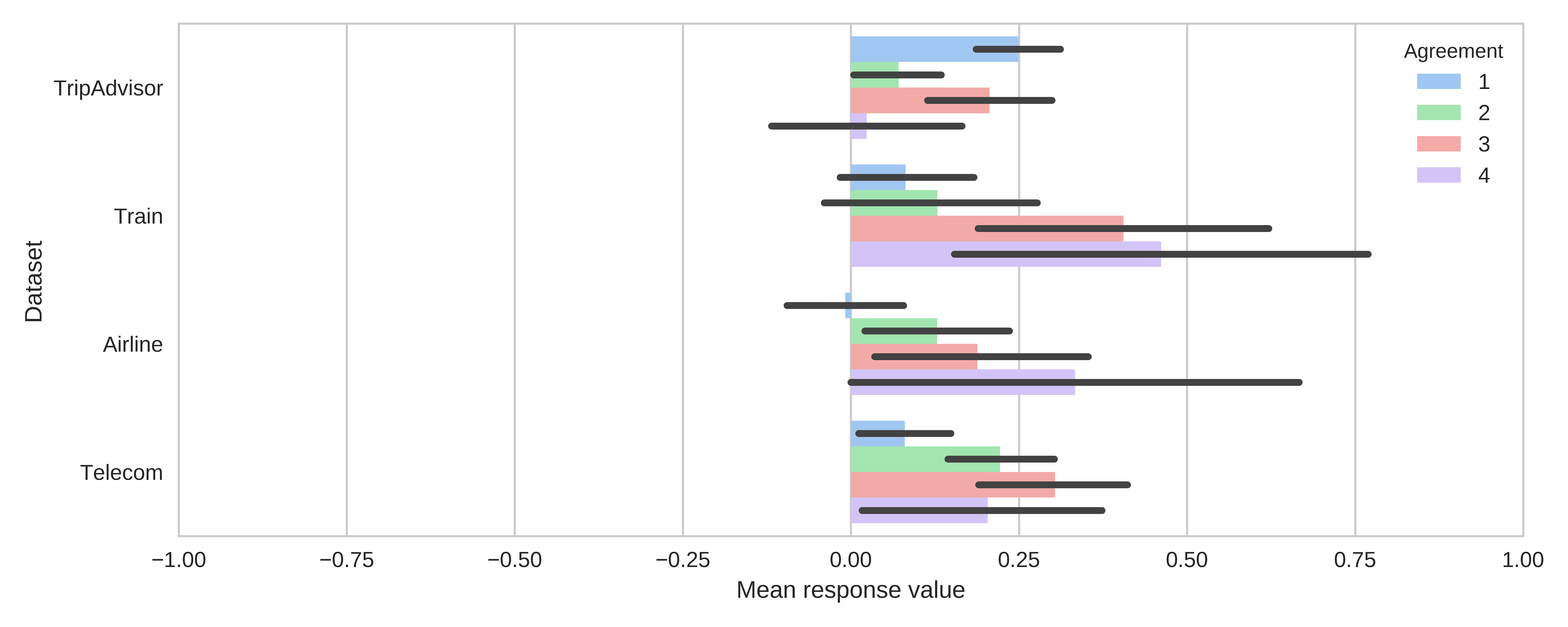}
\caption{Results of the A-B test on IVA response to original request versus cleaned request.  Black bars indicate 95\% confidence intervals.}
\label{fig:abtest}
\end{figure}

This A-B test was done only on responses that changed as a result of the cleaned request (3,588 IVA responses changed out of the 27,036 total responses). The result of this analysis is shown in Figure~\ref{fig:abtest}.  Note that the lower bound is -1, indicating the original IVA response is preferred.  If language is removed, the IVA response to the cleaned request is more likely preferred as made evident by the significantly positive skew.  95\% confidence intervals are included, and although they may seem large, this is expected; recall that a 0 was assigned if both IVA responses address the user request comparably or neither did.  In 10 of the 16 cases, the skew is towards the cleaned response within the 95\% confidence interval.

This is evidence that the current usage of unnecessary language has a measurable negative effect on live commercial IVAs.  TripAdvisor is an interesting exception, especially when the threshold is 4. However, this can be somewhat expected as it is a human-to-human forum where user inputs are significantly longer, and primary intent can be difficult to identify even for a human.

Although, in general, the removal of language is preferred, how \textit{much} removal? This is another question addressed in Figure~\ref{fig:abtest}. The higher the threshold, the more reviewers need to agree on the removal of the same segment of text. Thus, although language may still be removed, less language is removed with a high threshold than if the threshold was lower due to low kappa (see~\ref{subsec:agree}).  In effect, the higher thresholds may remove less unneeded language but the language that \textit{is} removed is more likely to be actually unnecessary which appears to improve the IVA understanding.  However, using a threshold of 4 seems to have limited improvement over 3 due to the reviewer disagreement.

\section{Conclusion}

Through the collection of this corpus and the annotation of relational segments, we have shown that users of commercial IVAs are already applying relational strategies to these IVAs. It is our prediction that these strategies will only increase as IVAs become more ubiquitous and human-like.  We have also shown that the removal of unnecessary language during intent determination not only increases intent classifier confidence but also improves response by reviewer standards.  It is our hope that by providing this data to the community, others can work on the automation of both the separation of business content from relational content and the classification of relational strategies.

The fact that it is possible to improve IVA responses to noisy forum data by the removal of relational language gives hope that automated methods of relational language detection may allow IVAs to contribute to human-to-human forum settings without substantial modifications to their underlying language models.  For example, an IVA could be employed on a commercial airline website while also monitoring and contributing to airline forum threads related to its company. This saves substantial effort and cost compared to deploying two special-purpose IVAs for each task.

Given the problematic presence of relational language in task-based inputs and our promising preliminary results, we encourage the research community to investigate ways on automating this annotation using our publicly available data\footnote{https://s3-us-west-2.amazonaws.com/nextit-public/rsics.html}.  There are many applications for such automation.  Determining if turns contain ranting in automatic quality assurance monitoring systems like the one presented by~\cite{roy2016qart} could help surface poor customer service more efficiently.  In systems for automating IVA refinement such as the one described by~\cite{beaver2016prioritization}, automatic detection of the presence of backstory or justification can be used as an indicator of possible missed intention.  In live IVAs, simplifying inputs before determining user intention as in our experiment can increase intent classification accuracy. Finally, IVAs can appear more human-like by classifying relational language to explicitly deal with relational content and respond in a relational manner.  Such applications would greatly improve the quality and relational abilities of customer service IVAs.

\appendix

\section{Annotation tools and process}

\begin{figure}[h]
\includegraphics[scale=0.9]{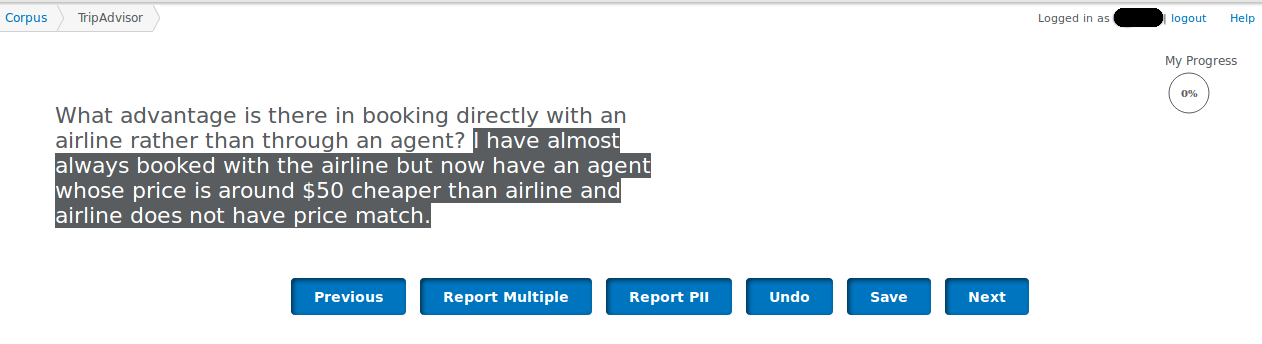}
\caption{Screenshot from the web-based annotation tool used by the eight reviewers for the first round of annotations.}
\label{fig:tool}
\end{figure}

The reviewers used a special purpose web application for the annotation tasks.  A screenshot of the interface used for the first task is shown in Figure~\ref{fig:tool}.  Each reviewer was required to complete a tutorial on tool usage and example selection tasks before they were allowed to start on the actual datasets.  A comprehensive explanation of the task was given to all reviewers before they began, and the authors were available by email to address questions or comments during the process.  As all reviewers were Next IT Corporation employees and knew the authors, there was ongoing communication through the task to ensure that annotations were correctly applied.  In addition, at any time they could click on the \textbf{Help} link at the top right to see the instructions given in Appendix B.

\begin{figure}[h]
\includegraphics[scale=.5]{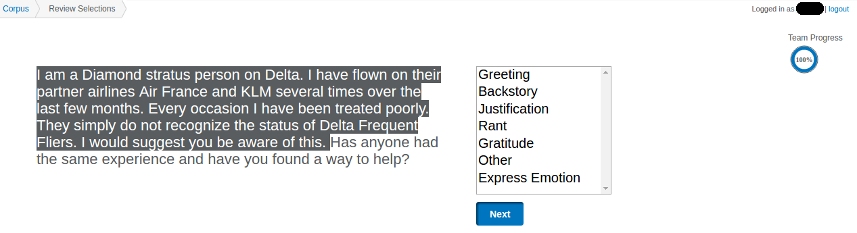}
\caption{Screenshot from the web-based annotation tool used by the authors for the tagging of relational segments.}
\label{fig:tool2}
\end{figure}

For the second round of relational tagging done by the authors, each highlighted section created by merging selections with an agreement of at least two reviewers was displayed.  An author then selected all relational tags that appeared within that highlighted section.  A screenshot of this interface is shown in in Figure~\ref{fig:tool2}.

\section{Instructions for Reviewers}
\label{apdx:help}

\textbf{Markup}

First, determine the purpose of the user query or statement.  Then, using the mouse, select \textit{all} subsections of the text that do not contribute to determining the purpose.  Only select text that is \textit{clearly} unneeded.  If more than one selection is required click the \textbf{Save} button between each selection.  If only one selection is required click the \textbf{Next} button to save the current selection and load the next text.  If no action is necessary click \textbf{Next} to load the next text.  If you need to return to the previous text click \textbf{Previous} (Note: this does not work if the last input was marked as Multiple Intent).  To remove a selection or edit a selection click \textbf{Undo} and reselect the intended text.

It is possible many texts will not need any markup.  Do not worry about including spaces on either side of your selection as they will be ignored.  Greetings and expressions of gratitude should always be marked as they are unnecessary to determining the intention, unless the entire text consists only of a greeting or thanks in which case they \textit{are} the intention.  If a text does not appear to have any clear intention or point no markup beyond greetings or emotion is needed.  In the following examples the gray text indicates selections that are unnecessary for determining the user's intent.

\noindent Examples:
\begin{itemize}
\item ``I will be traveling \textcolor{light-gray}{to see my husband before he leaves on a deployment }with my child under the age of 2 in March and I am looking for the cheapest price with her having a seat. \textcolor{light-gray}{Can you help.}"

\item ``\textcolor{light-gray}{Hi, }how can I change controls to allow R rated. \textcolor{light-gray}{I have no kids so I don't know why I don't have permission for this.}"

\item ``\textcolor{light-gray}{I did not get a reservation number.  }What number may I call to get my reservation number?"
\end{itemize}

\noindent\textbf{Multiple Intentions}

If an input \textit{clearly} contains two or more user intentions, click the \textbf{Report Multiple} button to flag it for removal and load the next text.

\noindent Examples:
\begin{itemize}
\item ``Hi, we are traveling with our Grandson and need to know what kind of identification we will need for him? Also, when we arrive in Penn station, will we be able to stop our one bag in the baggage area?"

\item ``hi I have a ticket \#\#\#\#\#\#  when must I book the ticket and when must I finish my trip. How much is the ticket worth? How much is the change fee?"
\end{itemize}

\noindent\textbf{Personal Identifiable Information (PII)}

If a text contains \textit{any} information that could be considered personally identifiable it needs to be flagged for cleanup.  PII includes usernames and proper names, credit card numbers, ticket numbers, confirmation numbers, phone numbers, account numbers, email addresses, zipcodes or mailing addresses, and any other information that could be used to identify an individual.

If you see anything you suspect of being sensitive click on the \textbf{Report PII} button and a red exclamation mark will appear next to the input.  It is better to report something you are unsure of than not report data that is actually sensitive.  However, there is no need to report data that is already sanitized such as ``My account number is xxx-xxxx".  After reporting the text, mark it up as usual and continue to the next user input.  All inputs reported with PII will be later reviewed and cleaned manually.

\begin{acknowledgements}
The authors would like to thank Next IT Corporation for sponsoring this work and sharing their data with the research community to further advance the human-like properties of IVA's.
\end{acknowledgements}




\bibliographystyle{authordate1}
\bibliography{paper}   

\end{document}